\documentclass[sigconf]{acmart}

\usepackage{enumitem}
\usepackage{comment}

\usepackage{placeins}
\usepackage{balance}

\AtBeginDocument{%
  }

\copyrightyear{2025}
\acmYear{2025}
\setcopyright{cc}
\setcctype{by-nc-nd}
\acmConference[KDD '25]{Proceedings of the 31st ACM SIGKDD Conference on
Knowledge Discovery and Data Mining V.1}{August 3--7, 2025}{Toronto, ON,
Canada}
\acmBooktitle{Proceedings of the 31st ACM SIGKDD Conference on Knowledge
Discovery and Data Mining V.1 (KDD '25), August 3--7, 2025, Toronto, ON,
Canada}
\acmDOI{10.1145/3690624.3709419}
\acmISBN{979-8-4007-1245-6/25/08}

\begin{document}


\title{Experimenting, Fast and Slow: Bayesian Optimization of Long-term Outcomes with Online Experiments
}


\author{Qing Feng}
\orcid{0009-0005-9535-4374}
\affiliation{%
  \institution{Meta}
  \city{Menlo Park}
  \state{CA}
  \country{USA}
}
\email{qingfeng@meta.com}

\author{Samuel Daulton}
\orcid{0000-0002-1448-7396}
\affiliation{%
  \institution{Meta}
  \city{Menlo Park}
  \state{CA}
  \country{USA}
}
\email{sdaulton@meta.com}

\author{Benjamin Letham}
\orcid{0009-0007-6480-9879}
\affiliation{%
  \institution{Meta}
   \city{Menlo Park}
  \state{CA}
 \country{USA}
}
\email{bletham@meta.com}

\author{Maximilian Balandat}
\orcid{0000-0002-8214-8935}
\affiliation{%
  \institution{Meta}
  \city{Menlo Park}
  \state{CA}
  \country{USA}
}
\email{balandat@meta.com}

\author{Eytan Bakshy}
\orcid{0009-0007-6480-9879}
\affiliation{%
  \institution{Meta}
  \city{Menlo Park}
  \state{CA}
 \country{USA}
}
\email{ebakshy@meta.com}

\renewcommand{\shortauthors}{Qing Feng, Samuel Daulton, Benjamin Letham, Maximilian Balandat, \& Eytan Bakshy}


\begin{abstract}
Online experiments in internet systems, also known as A/B tests, are used for a wide range of system tuning problems, such as optimizing recommender system ranking policies and learning adaptive streaming controllers. Decision-makers generally wish to optimize for long-term treatment effects of the system changes, which often requires running experiments for a long time as short-term measurements can be misleading due to  non-stationarity in treatment effects over time. The sequential experimentation strategies---which typically involve several iterations---can be prohibitively long in such cases. We describe a novel approach that combines fast experiments (e.g., biased experiments run only for a few hours or days) and/or offline proxies (e.g., off-policy evaluation) with long-running, slow experiments to perform sequential, Bayesian optimization over large action spaces in a short amount of time.
\end{abstract}



\maketitle


\section{Introduction}
\label{sec:intro}

Randomized online experiments are an important part of the product development cycle at internet firms and are used daily to evaluate product changes and drive improvements. The colloquial name \textit{A/B test} implies a comparison between two discrete variants A vs. B (or, control vs. test), but in reality online experiments frequently test several variants. Early work on causality in machine learning drew connections between A/B testing and the multi-armed bandit problem \citep{bottou2013counterfactual}, which has led to the term ``arm" being used to describe the different variants (test groups, configurations, etc.) under consideration. A typical flow is to evaluate the various arms via A/B tests, identify the best arm according to a collection of key metrics, and then deploy that arm as the new system default.

A/B tests are frequently used for complex tuning problems where the space of arms is continuous and multi-dimensional. For example, recommendation systems at large internet firms often sort content according to a ``value model'' or ``scoring function''~\citep{agarwal2018online} composed of continuous parameters (weights and offsets) specifying how various ML model predictions should be combined to form the ranking. As a simplified model, consider ranking an item with content features $\mathbf{c}$ for an individual with user features $\mathbf{u}$. A multi-task multi-label model produces $d$ predictions $f_i(\mathbf{u}, \mathbf{c})$ for the pair, which are combined with weights $x_i$ like: 
$s(\mathbf{u}, \mathbf{c}) = \sum_{i=1}^d x_i f_i(\mathbf{u}, \mathbf{c})$. The weights $\mathbf{x}$ form the space of possible scoring policies, and any point in that $d$-dimensional space can be an arm for A/B testing. Many other internet systems involve numerical parameters and thus require tuning with uncountable treatment spaces: streaming controllers, retrieval policies, ML inference platforms, and hyperparameter tuning in AutoML, to name a few.

\textit{Sequential model-based optimization} (SMBO) is a powerful approach for this class of tuning problems---that is, tuning in multi-dimensional spaces using a minimal number of evaluations, i.e., A/B tests. SMBO is an iterative process that uses response surface models to learn the relationship between the parameters being tuned and the metrics being measured. Model predictions are used to identify good arms for evaluation, those arms are evaluated in an A/B test, the model is updated with the results, and the process repeats iteratively. The use of modeling makes SMBO much more sample efficient than alternative derivative-free optimization approaches. However, it also makes SMBO inherently sequential: the results of previous rounds of A/B tests are necessary to fit the response surface model that directs the next round of tests. SMBO with a Bayesian response surface model, such as Gaussian process (GP) regression, is called Bayesian optimization (BO). Such models provide uncertainty estimates that are used to make an explore/exploit trade-off when selecting arms. BO has been applied successfully across a wide range of settings including simulation optimization \citep{santner2003design, roustant12, picheny13}, robotics \citep{Lizotte2007Automatic,Calandra2015a,Rai2018Bayesian}, AutoML \citep{hutter2011smac, snoek2012practical}, and, of course, tuning internet systems via A/B tests \citep{agarwal2018online, letham2019constrained, letham2019jmlr}. 

BO is used for online tuning across the internet industry, including at Meta \citep{fb_blog}, Google \citep{vizier}, Lyft, LinkedIn \citep{linkedin_blog}, and Pinterest. With this work, we hope to bridge some of the gaps between research and practice, and to describe for the first time the considerations of a deployed system that builds on the BO research happening across the field of machine learning. While the application of BO to A/B tests is established in industry, there is scarce literature describing how to do this effectively in practice.

\begin{figure}[tb]
    \centering
    \includegraphics[width=0.31\textwidth]{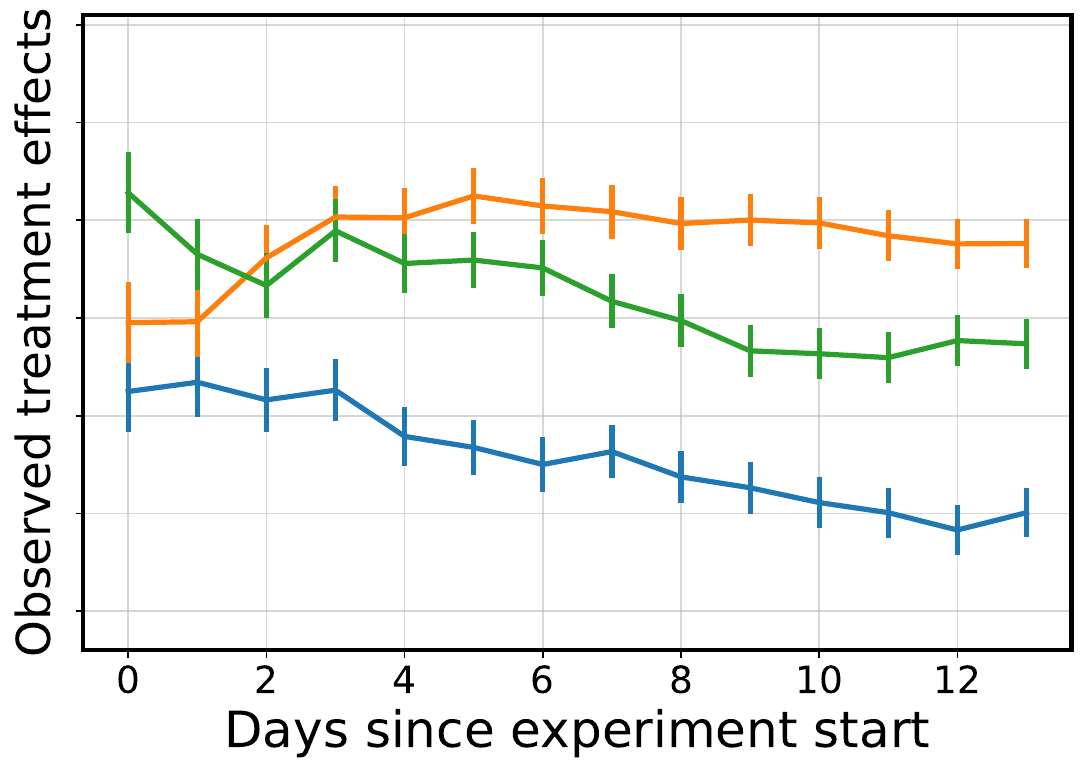}
    \vspace{-1.5ex}
    \caption{Treatment effect (mean and two standard errors) as a function of time for three arms in an information retrieval system tuning experiment. Nonstationarities in the treatment effect often necessitate running A/B tests for longer than two weeks.}
    \label{fig:treatment_effects}
    \vspace{-1ex}
\end{figure}

Our work focuses on a frequent pain point experienced by practitioners applying BO to A/B tests: targeting outcomes that require lengthy tests to measure. The goal of A/B testing is always to improve some set of metrics (i.e. outcomes), possibly subject to constraints on other metrics. While some outcomes like load times or stall rates can be measured over a short period, many important outcomes (usage, content production, interactions with others, etc.) involve complex and dynamic causal mechanisms, making a long-run A/B test the only way to reliably measure their treatment effects \citep{cheng2021long}. Fig.~\ref{fig:treatment_effects} shows an example of the time evolution of treatment effects for three arms in an A/B test tuning a retrieval system at Meta. In this experiment, the best arm (largest treatment effect) was different on day 1 than it was a week later, when the effects had reached a steady state.
Generally, relying on simple short-term proxies like the outcomes from the initial few days can lead to sub-optimal decisions~\citep{chen2019abtest,wu2022non,meta2022notifications}.

Long-running A/B tests pose a particular challenge to using sequential, adaptive methods like BO for tuning internet systems. Methods requiring three or more sequential rounds of A/B tests become impractical when each test requires several weeks. Past work has explored incorporating short-term proxies into models that expedite the decision-making process in A/B testing~\citep{athey2019surrogate, yang2023targeting}, however these approaches remain susceptible to bias. Proxy misspecification, lack of causal identification, and model miscalibration or staleness can all lead to poor optimization performance.

In this paper, we devise the \textit{fast and slow experimentation} framework, consisting of a novel experimental design and optimization methods that leverage proxies to directly target long-term outcomes with BO. Our main contributions are:
\begin{itemize}[leftmargin=16pt, labelwidth=!, labelindent=0pt]
    \item We introduce a parallel experimental design for efficient BO that combines short- and long-running A/B tests while targeting long-term outcomes. We show that this approach can work with experiments run online for as little as a few hours, and incorporate data sources at different time scales. 
    This experimentation and analysis flow 
    has been deployed in real-world experiments at Meta, achieving superior optimization results while reducing experimentation wall time by over 60\%.
    \item We introduce two models for learning long-term outcomes from (potentially biased) short-term proxies: multi-task GPs (MTGPs), and a novel \textit{target-aware GP} that can identify relevant short-term proxies from diverse information sources.
    \item We introduce a suite of benchmark problems emulating dynamic A/B testing environments observed in the real world, and with them show that our approach is able to effectively optimize for long-term outcomes.
\end{itemize}

\section{Background}

\subsection{Bayesian Optimization}
BO is a sequential strategy in which a probabilistic model is fit to the current set of observations and used to identify points in the parameter space to be evaluated next. The probabilistic model is usually a GP, which is a Bayesian nonparametric model that learns a smooth function $f(\mathbf{x})$ given the data. See \citep{rasmussen06} for a thorough introduction to GPs. GPs have great flexibility due to freedom in specifying the \textit{kernel}, $k(\cdot, \cdot)$, which determines the covariance between function values across the input space:
\begin{equation*}
    k(\mathbf{x}, \mathbf{x}'; \theta) := \textrm{Cov}[f(\mathbf{x}), f(\mathbf{x}')].
\end{equation*}
For instance, the classic radial basis function (RBF) kernel, $k(\mathbf{x}, \mathbf{x}'; \theta) = \textrm{exp}\left(-||\mathbf{x} - \mathbf{x}'||^2 / 2\theta^2\right)$,
can be used as the kernel in a GP model. The RBF kernel creates a spatially smooth model, since the covariance between function values decays as the distance between points grows. The kernel has hyperparameters $\theta$, such as the lengthscale in the RBF kernel; these hyperparameters are fit to the data by maximizing the marginal log likelihood. Much of the methodology in this paper will involve constructing new kernels that enable combining signals from different time scales of experimentation.

GPs have several properties that make them an excellent choice for BO. By virtue of being a Bayesian model, they provide not just a point estimate for the response surface, but also a full posterior distribution for $f(\mathbf{x})$. Moreover, that posterior is Gaussian, and its mean and variance can be computed analytically from the observed data. GPs are known to produce well-calibrated uncertainty estimates that are critical for decision-making under uncertainty.

Once the GP response surface model has been fit, the BO process continues by using an acquisition function $\alpha(\mathbf{x})$ to compute the value of selecting $\mathbf{x}$ as the next arm for evaluation. The acquisition function uses the model posterior for making this determination. Points will have high acquisition value due to some combination of high posterior variance (\textit{explore}) and high posterior mean (in a maximization problem; \textit{exploit}). Acquisition functions can also be evaluated on several points to construct a set of arms to be evaluated in parallel (a \emph{batch}). The arms with the highest acquisition value are selected for evaluation in the next A/B test, the GP is updated with the results, and iteration continues.

There is a large and growing literature on BO; see~\citep{frazier2018tutorial} for a review. Common choices for the acquisition function include Expected Improvement~\citep{jones98}, UCB~\citep{srinivas10}, Thompson sampling \citep{kandasamy2018parallelised}, and entropy search \citep{hernandez2014predictive,wang2017max,hvarfner2022joint}. Throughout this work we will consider single-objective optimization over a modest number of continuous parameters, but our methods are applicable a wide variety of BO problem settings (high-dimensional, multi-objective, combinatorial, etc.). Our focus here is on the models, which can subsequently be used with any of these popular acquisition functions.

Previous work has used cheap-but-biased proxies to aid optimizing with time-consuming evaluations through multi-fidelity BO~\cite{wu2020practical} or multi-information source BO~\cite{poloczek2017multi}. Other work has proposed multi-task BO with cost-sensitive acquisitions \citep{NIPS2013_f33ba15e} and access to offline simulations \citep{letham2019jmlr}. Our work leverages insights from that work to develop novel methodologies for addressing specific time-varying effects present in online experiments.

\subsection{Bayesian Optimization with A/B Tests}
\label{sec:bo_ab_tests}
Online systems are very amenable to randomized controlled trials since it is possible to create and deploy many variants of a system with minimal overhead. When individuals access the service, they are randomly assigned to one of the test arms, or to a control group. Individual-level outcomes are logged along with exposure to the treatment, so that aggregate outcomes can be measured and compared across arms. The result is an unbiased estimate of the treatment effect (i.e., the difference or ratio of the estimated effect of a test arm to a control arm) for each measured outcome, for each arm. A/B testing draws from the long history and literature on design and analysis of industrial and field experiments \citep{box2005statistics, gerber2012field}), as well as additional considerations specific to the setting of online experiments \citep{kohavi2009controlled, tang2010overlapping, kohavi2012trustworthy, bakshy2014designing}. 
The relative ease of deploying A/B tests belies significant challenges in using A/B tests to tune large numbers of parameters. BO provides a practical solution to these challenges, as we now describe.

\textit{Long durations.} The most conspicuous challenge to system tuning via A/B tests is the duration: Many tests take weeks to yield meaningful results due to changes in treatment effects over time and changes in the exposed population (more active users are more likely to be exposed early on) \citep{bakshy2014designing}. This time-consuming evaluation precludes the possibility of more than a few (2--4) rounds of iteration in practice. The high sample efficiency of BO, due to response surface modeling, enables finding good arms with limited iteration.

\textit{High parallelism.} Although the number of rounds of iteration is limited, one is usually able to test many arms in parallel. The common pattern for adaptive experimentation with A/B tests is to evaluate a large number of arms (e.g., 8--64) in parallel, depending on experimentation resource availability and on the signal-to-noise ratio of the outcomes. BO can construct batches of arms that are effective for optimization with high parallelism~\citep{ginsbourger2011, snoek2012practical, balandat19}.

\textit{High noise levels.} Even with millions of samples, noise levels in A/B tests are often much higher than what is commonly seen in other system tuning applications, such as hyperparameter optimization in AutoML. With a $1 / \sqrt{n}$ relationship between the sample size and the standard error, there are limited options for reducing the noise. However, this also produces a setting that matches the noise assumptions of GP models, and appropriate acquisition functions can handle these high noise levels \citep{letham2019constrained, daulton2021parallel}.

\textit{Smoothness.} Response surfaces in these settings are an average of outcomes over possibly millions of units, so a small change in, e.g., the value of specific reward signals in reinforcement learning systems, or the number of items retrieved from a recommendation source,
will generally produce a small change in the average outcomes. This creates a smooth response surface, consistent with the smoothness assumed by a GP. In practice we have found GPs to generally be an excellent model for A/B testing response surfaces. Furthermore, being an average of large numbers of observations means the measurements are in the CLT regime where the standard GP assumption of Gaussian noise is valid. 

\textit{Complex metric trade-offs.} Changes to online systems often result in complex trade-offs between several outcomes: improvements in computational efficiency may come with increased memory utilization, or one may need to balance different engagement metrics such as comments and video views. BO can be used with constrained~\citep{letham2019constrained} and multi-objective~\cite{daulton2020ehvi} acquisition functions in these cases.

BO has been the most successful tool for tuning online systems, at Meta and at other companies in the industry. BO is largely ``plug and play" with standard industry-scale A/B testing systems, since the only inputs that are needed are simple sufficient statistics produced by these systems. Reinforcement learning (RL) is a conceptually powerful framework for systems tuning, but it has not yet shown it can outperform BO. In the 2020 NeurIPS black box optimization challenge, none of the top methods used RL \citep{turner2021bo}. RL often requires thousands to millions of evaluations, as well as careful tuning (reward shaping) in order to be effective with multiple metrics. In fact, BO is the standard tool for tuning RL algorithms \citep{daulton2019thompson, mao2019videorl}. 
Classic bandit algorithms such as UCB or Thompson sampling have been used with GPs in the global optimization literature, though they are outperformed by more recent acquisition functions such as the qLogNEI that we use in our experiments here \citep{ament2023unexpected}.

As practitioners, our experience has been that the alternative to BO is expert manual tuning. The move to BO produces better results and improved intuitions about product trade-offs via flexible model-based analysis and visualization, all while freeing the experts to focus their attention on problems that are more interesting and less automatable than tweaking parameters.

\subsection{Estimating Long-Term Effects}
There are several reasons for long-running A/B tests, including novelty effects, learning effects, seasonality, data drift, etc.~\citep{bakshy2014designing,kohavi2020trustworthy}. Such behavior has been documented in a variety of contexts, including that of optimizing recommender systems at Meta~\citep{meta2022notifications}.
Multiple approaches have been proposed to accelerate estimation of long-term effects, including stratification and re-weighting to address exposure bias in A/B tests~\citep{kohavi2020trustworthy} or leveraging medium-term proxies to predict long-term effects~\citep{athey2019surrogateindex}. However, most work in this area focuses on estimating proxies that generalize across all experiments, and do not leverage specific information about the arms beyond their effects on intermediate outcome variables. In this work, on the other hand, we instead consider developing proxies that are tailored to each particular structured decision space (i.e., each arm is a $d$-dimensional vector of parameters). 
This approach is complementary to previous work on proxies: proxy metrics can be used as target outcomes, and provide a mechanism for leveraging short-term signals in a fashion that does not depend on the types of datasets or assumptions required by previous approaches (e.g., it only requires aggregate statistics from the experiment of interest).

\section{Experiment Design with Long- and Short-term measurements}

\begin{figure*}[t]
\centering
 \includegraphics[width=0.9\textwidth]{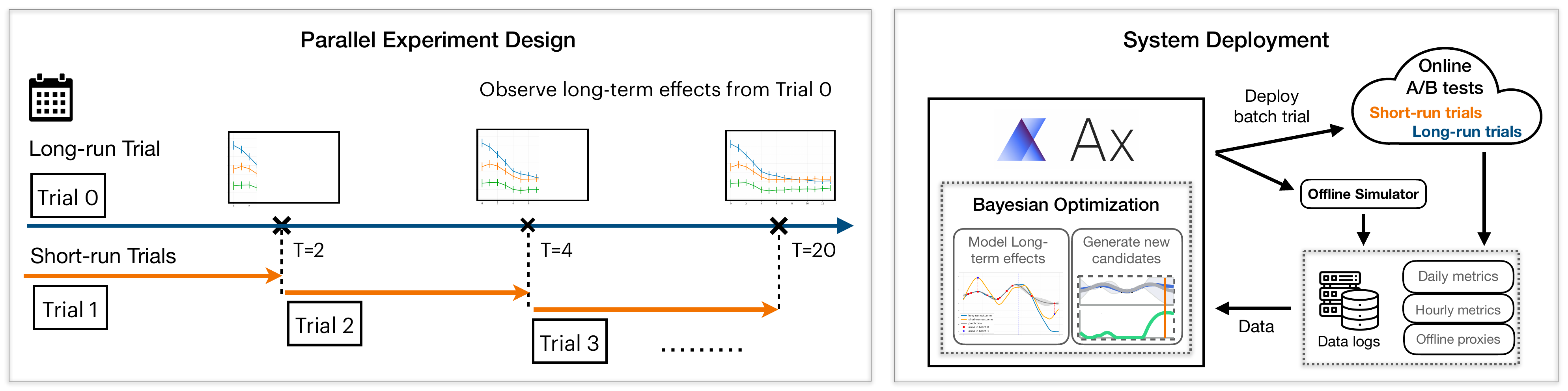}
 \vspace{-1ex}
\caption{Parallel experimental design for running fast Bayesian optimization. The long-run trial measures the long-term outcomes, while the short-run trials iteratively deploy candidates generated from optimization based on the predicted long-term outcomes, allowing for rapid exploration of the parameter space.}
\label{fig:flow}
\end{figure*}

\subsection{Parallel Experimental Design}
\label{sec:design}
To efficiently optimize long-term effects, we use a parallel experiment design, depicted in Fig.~\ref{fig:flow}. The experiment design uses a \emph{long-run experiment} (LRE) that measures the long-term effects (e.g. over 2 weeks) for a set of quasi-random arms sampled from the search space. In parallel with the LRE, a sequence of \emph{short-run experiments} (SREs) are run for much shorter periods of times (e.g. for a few days) in order to measure the short-term effects for arms that are selected via BO to target our best projection of the short-term outcomes to the long-term outcomes via the models described in Sections \ref{sec:model1}, \ref{sec:model2}, and \ref{sec:model3}. When prior knowledge is available (e.g. from previous related experiments \citep{perrone2019learning}), high-performing arms from prior experiments could be for warm-starting the first SRE, and otherwise quasi-random arms are used.

We deploy this experiment design by running two mutually exclusive experiments (LRE and SRE) in parallel. At the start of each experiment (the LRE and each SRE), a subset of users are randomly sampled from the population and are randomly allocated to different treatment groups. 
Every few days, the current SRE is stopped and replaced with a new SRE. We leverage the Ax software (available in open source at \url{http://ax.dev}) to directly interfaces with the experimentation platform to allow generating new arms via BO and deploying new SREs via Python APIs.

The entire optimization is completed in the same time as running a single long-run experimental trial, and is thus $K$ times faster than a traditional sequential design with the same $K$ trials ($K=4$ is common in our deployments).
This parallel design allows us to learn long-term time-varying effects at different points across the parameter space, while identifying promising regions of the parameter space through SREs. The SREs can also be used to transfer knowledge about kernel hyperparameters, enabling the model to more efficiently learn from limited long-term outcomes (for additional discussion on multi-task generalization with MTGPs, see~\citep[Sec. 6]{letham2019jmlr}). 
In addition to time efficiency, the new design  improves the total experiment resource efficiency (allocation per day $\times$ run time). Given the same allocation of experimental resources, we have observed, both in practice and via reproducible simulation studies, that the \emph{fast and slow} approach saves significant experimentation resources--over 60\%--compared with a traditional one in which a limited number of long-running sequential batches are deployed.

\subsection{Deployment}
As Fig.~\ref{fig:flow} illustrates, at each iteration, metrics are fetched and new arms are generated using BO. The experiment is managed by an internal service built on top of Ax, Meta's open-source platform for adaptive experimentation, and deployed via the company's internal A/B testing system. When the SREs are relatively long (e.g. a few days), a human-in-the-loop approach is often used when selecting the next batch, so that trade-offs and objectives are reconsidered for each time-consuming batch. With sub-daily trials, experiment iteration is fully-automated, requiring no human intervention. Thanks to this automation, from a user perspective running \emph{fast and slow} experiments is only slightly more involved than running typical parameter tuning A/B tests.

Our experimentation framework has been widely deployed across the company, achieving significant successes in various products. On average, it has produced more than 30 major launches per year, which have substantially contributed to teams’ top-line goals and accounted for a significant portion of team biannual targets. For example, we applied our methods to optimize retrieval systems in ranking, resulting in more efficient use of computational resources. This effort alleviated capacity crunches, saving over 2 MW of capacity, while simultaneously improving ranking performance.

\subsection{Practical Considerations}
A challenge in applying adaptive experimentation to industry experiments is that decision-makers often consider many trade-offs when making launch decisions, and it can be difficult to translate these requirements into an optimization objective. 
A common pitfall faced by first-time users of the system is to not specify all of the necessary outcome constraints (sometimes referred to as a \emph{counter-metrics}). Optimizing the target metric can then negatively impact other metrics.  For example, an engineer wishing to maximize video upload quality may end up in a situation in which all videos are uploaded at high quality, regardless of the quality of people's internet connections, resulting in poor upload reliability \citep{daulton2019thompson}.
Education and direct engagement with users of our system (often via talks, our internal Q\&A group, office hours, or direct collaboration) help reduce these challenges, as have integration of better methods and interfaces for navigating multiple objectives~\citep{daulton2021parallel, pmlr-v202-daulton23a, lin2020preference,lin2022preference}.

Engineers may also select metrics that are causally far downstream from the specific intervention they are making, which can unnecessarily increase variance.  For instance, an engineer working to improve a cache eviction policy on a mobile device may attempt to directly optimize for daily active users. However, the parameters being tuned most immediately effect much easier-to-measure metric movements on short-term effects, such as cache misses and hits, data utilization, and memory consumption.

Carry-over effects, where past interventions influence current outcomes, are potential practical issue when running rapid SREs. To address this, we randomly allocate individuals to groups at the start of the experiment, and then those assignments are re-randomized at each iteration. This ensures that, over time, each arm experiences similar carry-over effects, contributing to variance in subsequent actions but avoiding bias from carry-over effects.

\section{Models of Long-term Outcomes using Short-term Information}\label{sec:model1}
In the following sections we discuss three modeling approaches utilizing the \emph{fast and slow} design.
In this section, we describe a multi-task model for when there is a one-to-one mapping between short-term proxies and long-term targets. Section \ref{sec:model2} introduces a new target-aware GP model, useful for a many-to-one mapping between proxy and target metrics. Finally, Section \ref{sec:model3} describes a model for handling time-of-day effects with hourly measurements. 

We first consider the case in which outcomes from the LRE have some corresponding metric or proxy that can be derived from the SREs. For instance, we can use a 2-day measurement of a target engagement metric as a proxy for the 2-week long-term measurement of that same metric.
With multi-task modeling, these short-term measurements are able to rapidly constrain the search space and are informative about which inputs impact the metrics the most.

We leverage an MTGP to learn how the outcomes vary across the parameter space and time, while continuing to preserve uncertainty quantification needed for performing BO.
Given one LRE and $S$ SREs, the MTGP jointly models the outcomes from that collection of functions $\{f_L(\mathbf{x})\} \cup \{f_s(\mathbf{x})\}_{s=1}^{S}$.
Here, $f_L$ represents the long-term outcome, and $f_s$ the outcome of SRE $s$. The MTGP is constructed by designing a kernel that models the covariance across both trials and the parameter space.
That is, the kernel is specified as $k((t, \mathbf{x}),(t', \mathbf{x'})) = \textrm{cov}[f_t(\mathbf{x}), f_{t'}(\mathbf{x'})]$, in which $t$, $t'$ are the task indices corresponding to any pair of trials from $\{L, 1, \cdots, S\}$. A common choice for a multi-task kernel is the intrinsic coregionalization model (ICM) kernel \cite{bonilla2008multi}, which assumes separability between the covariance across trials and across parameters:
\begin{equation*}
    k((t, \mathbf{x}),(t', \mathbf{x'})) = K_{t, t'}\,k^{x}(\mathbf{x}, \mathbf{x'}; \theta).
\end{equation*}
Here $K$ is an $(S+1) \times (S+1)$ positive semi-definite matrix that models the correlations between trials (the task covariance matrix), and $k^{x}(\mathbf{x}, \mathbf{x'}; \theta)$ is the kernel over the parameter space, e.g. the same kernel as in a regular GP. A free-form task covariance matrix $K$ can be estimated alongside the spatial kernel hyperparameters $\theta$ by maximizing marginal log likelihood~\cite{bonilla2008multi}. The full details of the model implementation can be seen in the software release.

\begin{figure}[tb]
\centering
 \includegraphics[width=0.45\textwidth]{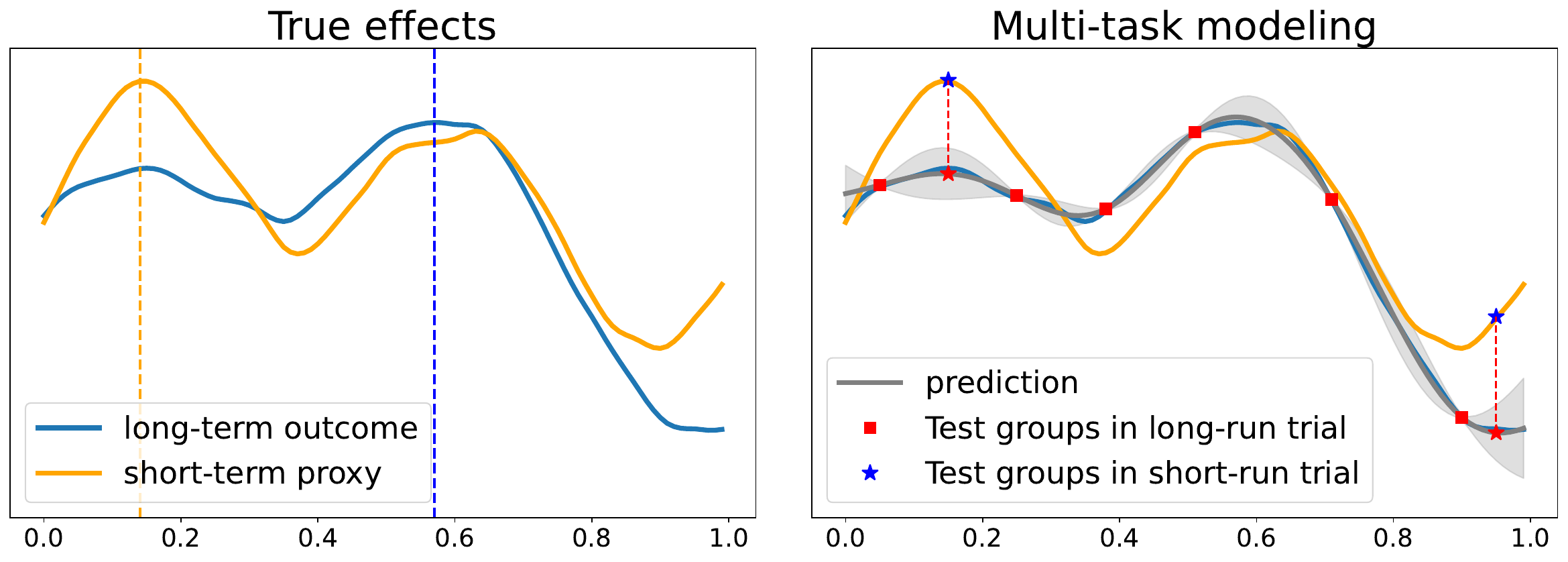}
\caption{Using an MTGP to learn the effects over time and the parameter space. \emph{Left:} long-term and short-term outcomes. Dashed vertical lines show that the optimum for the long-term outcome ($\approx$0.6) is different than that of the short-term outcome ($\approx$0.15).  \emph{Right:} long-term (red squares) and short-term (blue stars) outcomes are observed from the long- and short-run experiments. The MTGP accurately predicts the long-term effects (grey curve), despite not observing long-term outcomes for the short-run experiment.}
\label{fig:mtgp}
\end{figure}

\begin{figure}[htb]
\centering
 \includegraphics[width=0.475\textwidth]{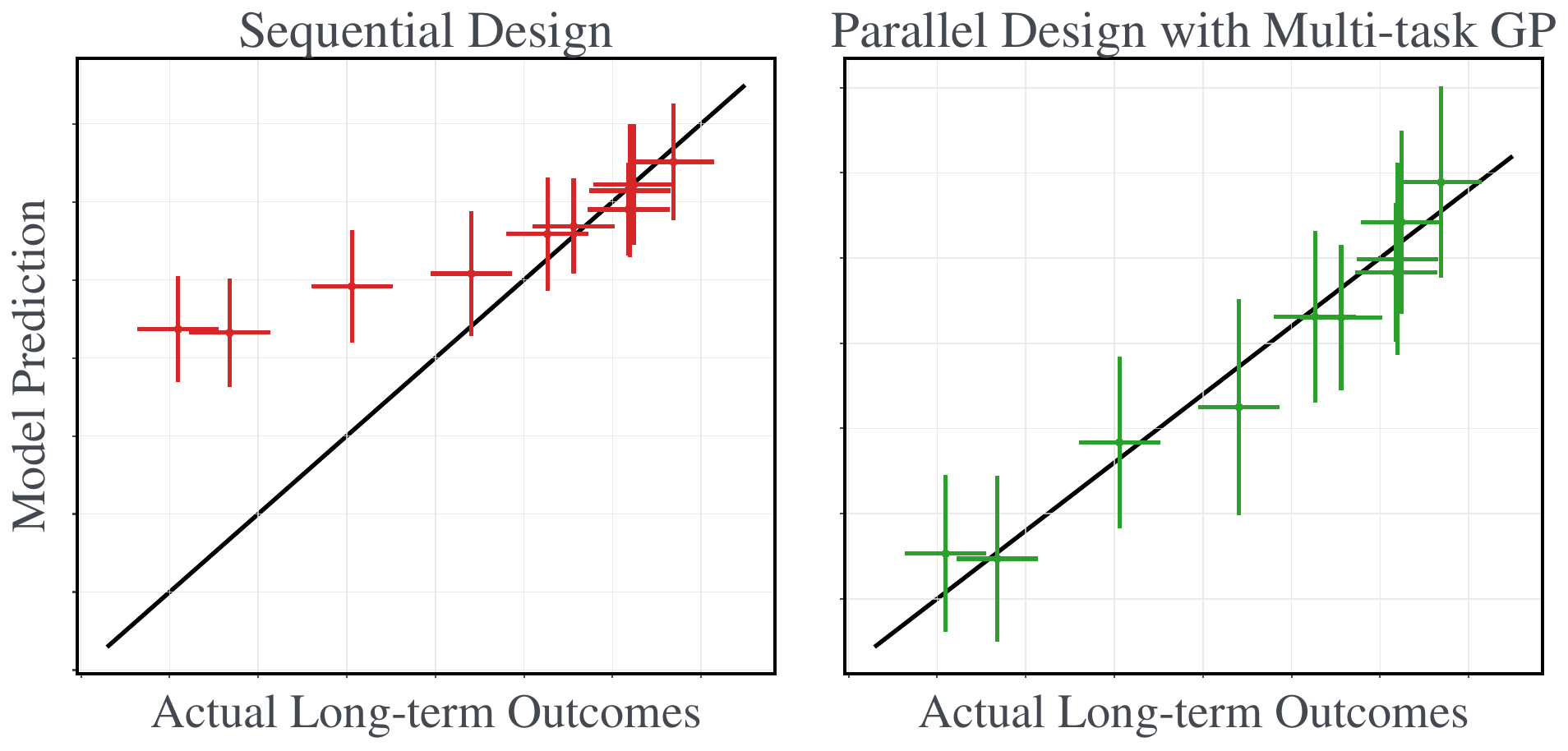}
\caption{Leave-one-out cross validation predictions for GP models fit to a real experiment. (Left) Predictions of a GP using only short-term outcomes are strongly biased. (Right) MTGP predictions show much better fit. Horizontal lines indicate 95\% confidence intervals of the observed mean and vertical lines indicate 95\% posterior predictive intervals.}
\label{fig:cv_long_term_mtgp}
\end{figure}

By learning correlations between trials, the biased short-term observations can be used to explore the parameter space and accelerate the optimization; observations from the LRE correct the bias of the short-term measurements, allowing one to accurately predict the long-term outcomes. Fig.~\ref{fig:mtgp} illustrates the MTGP with a 1-d toy example. 
Fig.~\ref{fig:cv_long_term_mtgp} presents the leave-one-out cross validation predictions of a long-term metric from a real experiment. The results reveal that the parallel design with an MTGP enables accurate prediction of the long-term outcome, whereas a sequential design with short-term metrics or proxies alone can introduce significant bias. 
Using the GP posterior over the long-term outcomes, we generate new arms to evaluate in the next short-run trial. 

\section{Leveraging
Proxy Metrics}\label{sec:model2}
When modeling the long-term behavior of experiments, there are often numerous causally upstream variables that are correlated with long-term outcomes of interest, and which can help improve the model. This can include data from short online experiments, as well as high-throughput offline evaluations.

\subsection{Online and Offline Proxy Metrics}

Typically a large number of online metrics take the form of de-identified aggregate statistics. Such rich aggregate information can capture various facets that underlie the causal mechanisms behind the interventions being tested. Notably, these metrics are often correlated with each other. For example, short-term movement in metrics like click-through rates and rates of re-sharing a post may give useful insights into engagement patterns over a longer period of time. These provide one useful source of proxy metrics.

A second source of proxy metrics can be an offline simulator, such as those employing off-policy evaluation~\citep{bottou2013counterfactual}. Simulators are an attractive source because they provide significantly higher evaluation throughput than online A/B testing. While off-policy evaluation in reinforcement learning remains a challenging task \citep{thomas2016data, munos2016safe}, constructing simple offline simulators for recommendation systems can considerably expedite the tuning process~\cite{letham2019jmlr}. Despite the bias inherent in offline estimates, as too with online short-term proxies, they often provide valuable directional information.

\begin{figure}[t]
\centering
 \includegraphics[width=0.32\textwidth]{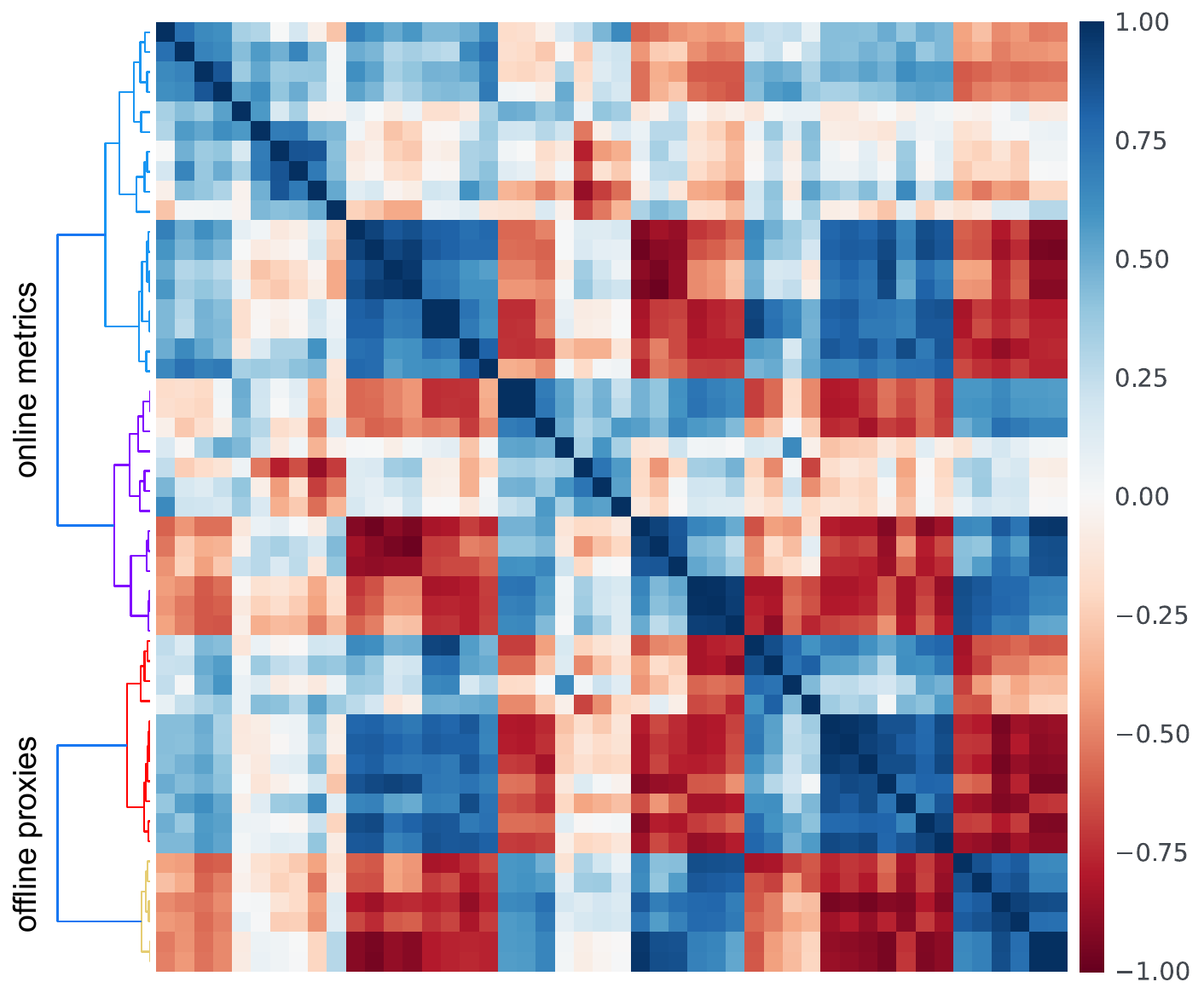}
 \vspace{-1.5ex}
\caption{Correlation between metrics including online metrics and offline proxies obtained from a simulator. Each group (online and offline) was separately clustered, producing the two dendrograms on the left axis used to order metrics. The top block are online metrics, the bottom block are offline proxies. There are strong correlations between online metrics, and between offline proxies and online metrics.}
\label{fig:metric_correlation}
\vspace{-1ex}
\end{figure}

Fig.~\ref{fig:metric_correlation} shows a heatmap of the correlation between metrics, across a set of both online metrics from a recommender A/B test, and offline proxies from a simulator of the recommendation system. Clear correlation structures are observable within each data source, as well as correlations across online and offline data sources. This demonstrates the value of borrowing strength between online metrics, and of incorporating offline proxies into online tuning.

With many possible proxies for a metric of interest, identifying \textit{which} proxies are truly informative is challenging. This makes it difficult to use the multi-task model described in Section \ref{sec:model1}.
Na\"ively combining all proxies in an MTGP may lead to negative transfer: including poor proxies can cause the ICM MTGP to perform worse than using no proxies at all. This happens if tasks require different lengthscale hyperparameters (e.g., due to different feature importances), or are not well-correlated with each other.
In order to make use of many proxies, we develop a new multi-task model that distills relevant information while guarding against negative transfer.

\subsection{Target-Aware Gaussian Process Model}
As before, we denote the target long-term outcome as $f_L(\mathbf{x})$ and the various short-term proxy metrics, both online and offline, as ${f_s(x), s= 1, \cdots, S}$.
The target-aware GP (TAGP) model comprises two main components: a set of pre-trained \emph{base models} that have been fit to the proxy metrics, and an ensemble model that combines the sources to predict the target metric.
A separate (single-task) GP is fit to the data from each proxy, for $S$ base models. In contrast to the MTGP of Section \ref{sec:model1}, where all tasks shared a spatial kernel with hyperparameters $\theta$, here each model has its own kernel with independently fit hyperparameters $\theta^s$, $s=1, \ldots, S$. This is key for avoiding negative transfer across a large set of proxies, as kernel mismatch across tasks is a significant source of negative transfer.
The target long-term outcome is then modeled as a weighted average of a bias term $g(\cdot)$, also a GP with independent kernel, and the GP posterior of each base model:
\begin{equation*}
f_L(\mathbf{x}) = g(\mathbf{x}) + \sum_{s=1}^S w_sf_s(\mathbf{x}).
\end{equation*}
The kernel hyperparameters of $g$ and the base model weights $w_s$ are fit by maximizing the leave-one-out cross-validation likelihood for the target long-term outcome~\citep{rasmussen06}. As $g$ and each $f_s$ has a Gaussian posterior, $f_L$ will have a posterior that is a mixture of Gaussians. 

This approach relaxes the assumption of sharing the same spatial kernel across all sources made by the ICM kernel in Section \ref{sec:model1}, as each proxy now has its own kernel function. The target-aware model also has scalability advantages, easily accommodating cases with many proxies. Computation scales linearly with the number of base models $S$, whereas with the ICM kernel it scales cubically and thus can become unwieldy even for small $S$.

The base model weights $w_s$ are regularized with a half-Cauchy prior. This regularization enables the model to shrink the weights of irrelevant information sources toward zero, which is also important for mitigating negative transfer. The imposed sparsity enables the target-aware model to selectively choose useful proxies guided by the target long-term outcome. In the extreme where all weights $w_s$ are shrunk to zero, the target-aware model reverts to a standard single-task GP trained solely on the target data.

\begin{figure}[t]
\centering
 \includegraphics[width=0.475\textwidth]{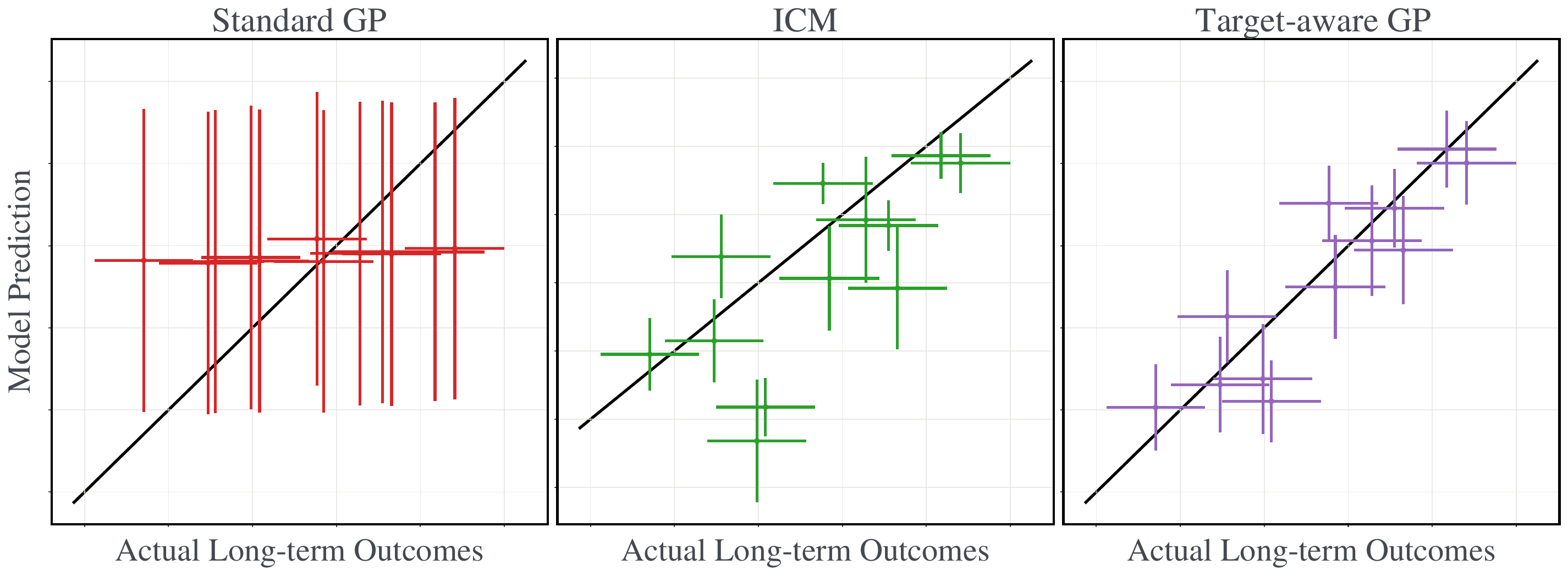}
 \vspace{-1.5ex}
\caption{Leave-one-out cross-validation predictions from a real experiment, highlighting the effectiveness of the target-aware GP in distilling information from multiple proxies to accurately predict long-term outcomes.}
\label{fig:cv_long_term_mtgp_tagp}
\vspace{-1ex}
\end{figure}

Fig.~\ref{fig:cv_long_term_mtgp_tagp} shows the leave-one-out cross-validation predictions of three GP models fit to data from a real online experiment at Meta. The standard GP, trained exclusively on long-term outcomes, tends to predict the mean with high predictive uncertainty. The MTGP model with an ICM kernel shows negative transfer. In contrast, the target-aware GP demonstrates its capacity to distill information from biased proxies, effectively reducing uncertainty while maintaining robustness against negative transfer. 

\section{Short-term Proxies with Time-of-Day Effects}\label{sec:model3}

\begin{figure}[tb]
    \centering
    \includegraphics[width=0.45\textwidth]{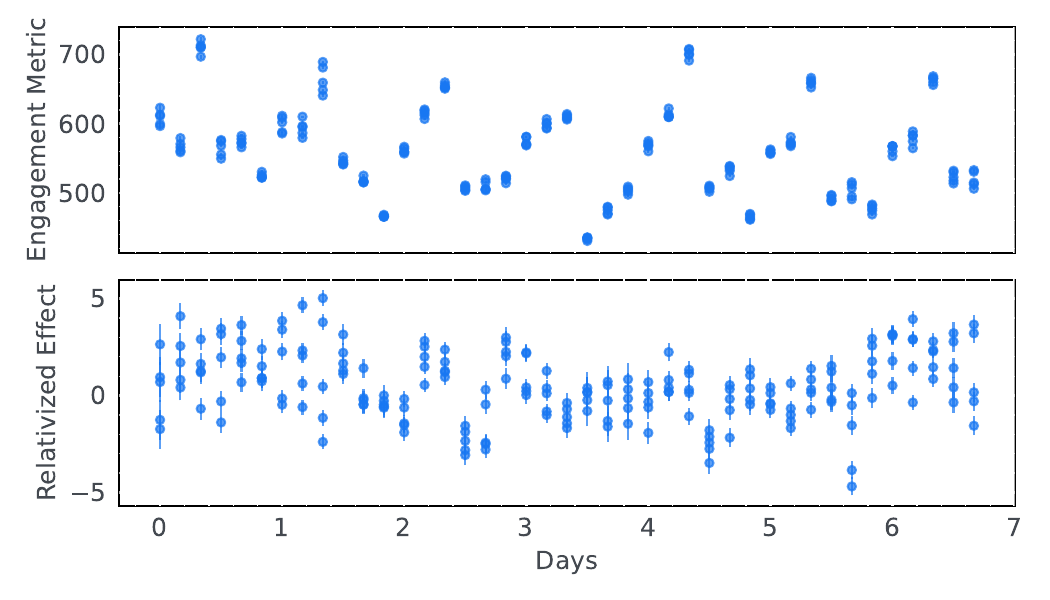}
    \vspace{-2.5ex}
    \caption{\label{fig:temporal_effects} Metrics and treatment effects from an online experiment exhibiting periodic trends. Each point is an observation of one arm. New arms are used at each iteration (every 4 hours). The top panel shows periodicity in the metric itself. The bottom panel contains the respective treatment effects (relativized w.r.t. the control), which also display substantial levels of non-stationarity.
    }
    \vspace{-1ex}
\end{figure}

Our initial deployment of the \textit{fast and slow} design used 2--3 days for SREs, for maximum compatibility with internal experimentation platform tooling. The next frontier of accelerating online tuning is to reduce the SREs from a few days to a few hours. Running A/B tests with timescales of hours introduces significant time-of-day effects, as traffic and product usage fluctuate significantly throughout each day. These periodic effects are visible in the observations in Fig.~\ref{fig:temporal_effects} collected during a real-world tuning experiment for a recommendation system. Notably, even after partially controlling for time-of-day effects by relativizing the metrics with respect to the control arm, there is still significant non-stationarity in the treatment effects (Fig.~\ref{fig:temporal_effects}, bottom), suggesting that there is treatment effect heterogeneity with respect to time.

We can significantly improve the quality of our predictive models by accounting for these time-of-day effects. We do this by incorporating time (hours elapsed since the start of the experiment) as a feature in the GP kernel. That is, we model each metric as a function $f(\mathbf{x}, \tau)$ of the arm parameters $\mathbf{x}$ and time $\tau \in \{0,1, ..., T\}$. This is done by introducing a \textit{temporal} kernel:
\begin{equation*}
    \label{eqn:temporal_kernel}
    k_\mathcal{T}((x,\tau), (x',\tau')) = \sigma_f k^\text{P}(\tau,\tau'; p, l_p) k^{x}(\mathbf{x}, \mathbf{x'}; \theta).
\end{equation*}
Like the ICM kernel of Sec. \ref{sec:model1}, this kernel is a product of two kernels, in this case a kernel over time $k^\text{P}$ and the usual spatial kernel $k^{x}$. The term $\sigma_f$ is a signal variance hyperparameter that is fit along with other kernel hyperparameters, and the kernel over time is:
\begin{equation*}
    k^{\text{P}}(\tau, \tau') = \exp \left(- 2 \sin^2 \left(\pi|\tau-\tau'|/ p \right) /l_p^2\right).
\end{equation*}
This kernel has as hyperparameters $p$ (period length) and $l_p$ (lengthscale). The period length for time-of-day effects is known and fixed to $p=24$ hours (additional periodic effects, such as day-of-week, could be modeled using an additional temporal kernel with different periodicity). All other model hyperparameters are jointly fit by maximizing the marginal log likelihood in the usual way.

The key assumption of the temporal kernel is that time-of-day effects are periodic. This assumption is validated empirically in Fig.~\ref{fig:temporal_effects}, making the kernel a suitable choice. In real experiments in which this model was deployed, accounting for time of day improved log-likelihood in 12 of 14 engagement metrics and improved mean squared error in 6 of the engagement metrics (see Table~\ref{table:time_aware_model_diag} in the Appendix), resulting in improved predictions for use in BO.

Leveraging the \textit{fast and slow} experiment design described in Section~\ref{sec:design}, we keep one trial fixed to measure long-term effects and use a sequence of rapidly evolving short-run trials, where new trials are deployed every few hours in a fully-automated fashion. We leverage the MTGP of Section \ref{sec:model1} to model the inter-task correlations across the model-estimated time-averaged short-term effects for the short-run trials (first task) and the long-term effects for the long-run trial (second task). The MTGP is used to select best arms that optimize the long-term effects, and we run those arms in a final trial to measure the long-term effects.

\section{Experiments}

\begin{figure*}[t]
    \centering
    \includegraphics[width=0.95\textwidth]{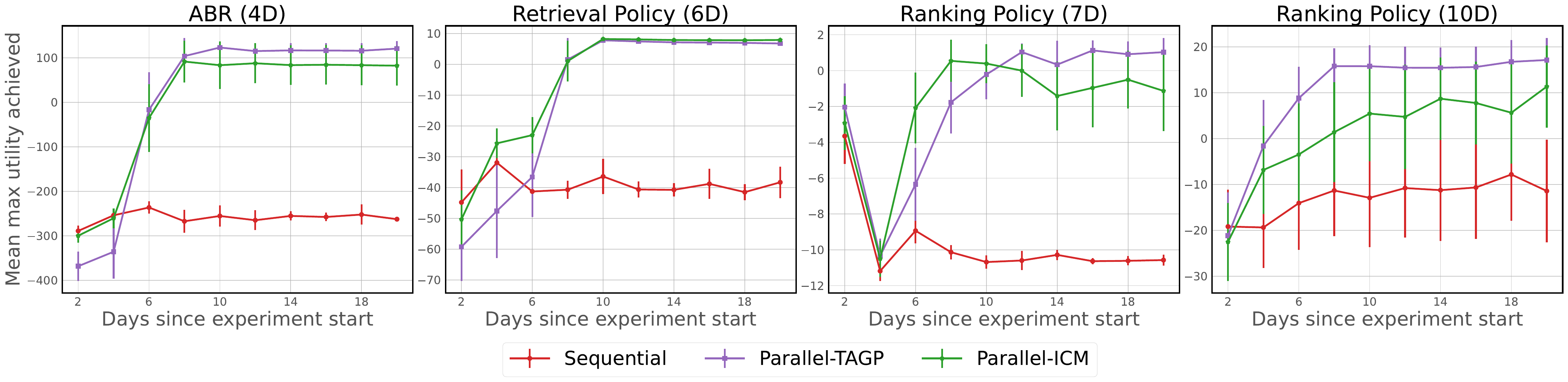}
    \vspace{-1.0ex}
    \caption{\label{fig:benchmark} Benchmark results from four real-world simulation studies showing the significant improvements achieved in the optimization of long-term outcomes using our parallel experimentation design and appropriate multi-task models.}
\end{figure*}

We now present empirical results from our \textit{fast and slow experimentation} framework showing it can effectively optimize long-term outcomes. Our evaluations include both synthetic problems and real-world-based simulation studies across diverse applications, such as video playback controller  optimization and retrieval and ranking policy optimization for recommender systems. All methods are implemented using BoTorch \citep{balandat19} and Ax \citep{bakshy2018ae}. Code for the models is available at \url{https://github.com/facebookresearch/FastBO}.
For all experiments, all methods use the qLogNEI acquisition function \citep{ament2023unexpected}, an improved variant of noisy EI \citep{letham2019constrained}, which is designed for the high noise levels and constraints of the A/B test setting. The Mat\'ern kernel was used for the spatial component of each kernel.

To evaluate the performance of each method in our benchmarks, at each iteration we identify the (possibly out-of-sample) arm with the best posterior mean. Methods are evaluated by the ground-truth value of this model-identified best arm. This reflects how decisions are made using the model in practice, where true long-term outcomes are not observed.

\subsection{Experiments Without Time-of-Day Effects}
We evaluate sequential and parallel (\emph{fast and slow}) experiment designs,
together with the proposed GP models under different data source scenarios. To ensure a fair comparison, the total experiment resources, defined by the number of arms running concurrently and the overall wall time, are fixed for both designs. In the first four experiments, each iteration spans a duration of 2 days, with a total experiment duration set at 20 days.
Arms are equally divided between the LRE and SREs. We compare the optimization performance of:
\begin{itemize}[leftmargin=16pt, labelwidth=!, labelindent=0pt]
    \item \textit{Sequential}: A standard GP using SREs only.
    \item \textit{Parallel-ICM}: Jointly modeling short- and long-term outcomes with the parallel design and MTGP (Sec. \ref{sec:model1}).
    \item \textit{Parallel-TAGP}: The parallel design with the target-aware GP model (Sec. \ref{sec:model2}) incorporating \textit{all} short-term outcomes in the model for each long-term outcome.
\end{itemize}

We consider 4 benchmark tasks based on diverse real-world policy optimization problems. The first two tasks are modifications of existing benchmarks, while the latter two are based off de-identified aggregate statistics from online experiments conducted at Meta.  Additional simulation problems can be found in the Appendix.  To emulate time-varying effects found in the real-world, we apply a time-varying effect transformation the values at each arm, that depends on how long the experiment has been running and the parameter values. This simulates real-life settings where each outcome converges to its true long-term value over time, and different arms experience different time-varying effects. Additional details are included in Appendix~\ref{app:tv_effect}. 
All figures show the mean and 2 standard errors of the mean across 25 replications, and additional details for each of the problems below are in the Appendix.

\textit{Video Playback Controller.} Video streaming and real-time communication systems rely on adaptive bitrate (ABR) algorithms to balance video quality and uninterrupted playback. We use the contextual ABR controller policy problem introduced in~\citep{feng2020cbo}, which was derived from real, de-identified video playback trace data from the Facebook Android mobile app. We optimized 4 parameters of the ABR controller policy in order to maximize quality while minimizing stall time. Auxiliary metrics such as stall time and video chunk quality were used as additional proxies for the TAGP.

\textit{Retrieval Policy.} The sourcing component of a recommendation system collects a set of items sent to the ranking algorithm for sorting. This process involves retrieving items from multiple sources, each potentially representing distinct facets of the user interest taxonomy~\cite{wilhelm2018dpp}. Querying for more items may enhance the recommendation system's quality, at the cost of increasing infrastructure load. We use the recommender sourcing system simulation developed by~\citep{liu2023sparse}. We optimized over 6 parameters, each specifying the number of items to be retrieved from six selected sources, to maximize content relevance and minimize infrastructure load.

\textit{Ranking Policy.}
This problem used a pair of surrogate functions fit to data collected from a series of online experiments tuning the value model (described in the Introduction) of Instagram's Feed recommendation system.  
The ranking policies tuned here were parameterized by 7-d and 10-d value model parameter sets. In both cases we maximize engagement metrics, while observing various additional metrics of ranking quality. 

Fig. \ref{fig:benchmark} show that the parallel design significantly improves on a standard sequential design, and that tuning can be further accelerated by incorporating additional proxies via the TAGP model.

\begin{figure}[t]
    \centering
    \includegraphics[width=0.35\textwidth]{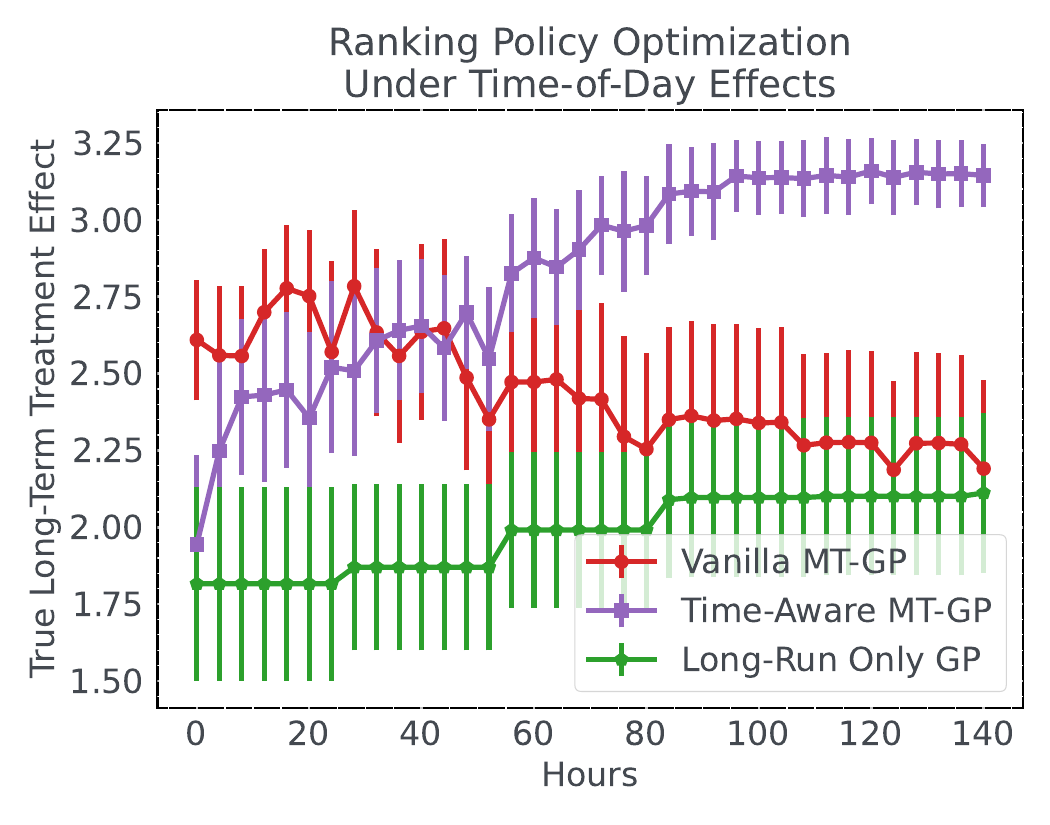}
    \vspace{-1.5ex}
    \caption{\label{fig:realtime_benchmark} Results on optimizing a ranking system with short-run evaluations that are subject to time-of-day effects.}
\end{figure}

\subsection{Experiments with Time-of-Day Effects}
\textit{10-d Ranking Policy Optimization.}
We used the 10-d ranking policy problem from above to test optimization with time-of-day effects. Here we incorporated faster SREs, with SREs of 3 arms each running for only 4 hours, alongisde an LRE of 6 arms.  We evaluate three BO methods: 1) ICM MTGP ignoring the time of day effects (Sec. \ref{sec:model1}); 2) a time-aware MTGP with the temporal kernel to account for the time of day effects (Sec. \ref{sec:model3}); and 3) a standard GP fit just to the LRE.

Fig.~\ref{fig:realtime_benchmark} shows that accounting for time-of-day effects with the temporal kernel significantly improves optimization performance. The temporal kernel will be most useful when the time-of-day effects are large relative to the overall signal, and when evaluations are short (e.g. a few hours).

\section{Discussion}
BO is a powerful tool for optimization of online systems. These methods are complementary to techniques in the A/B testing literature such as variance reduction and proper handling of interference \citep{deng2013cuped,karrer2021network}. In this paper, we propose a \emph{fast and slow experimentation} framework to efficiently perform BO with A/B tests.
Our parallel design enables rapid exploration of the parameter space, and effectively makes use of short-run experiments with time scales as short as hours.
While the simulation studies shown here consider low-dimensional single-objective problems for the sake of presentation, the approach is also deployed to optimize high-dimensional, multi-objective problems in large-scale recommender systems at Meta. There are many areas for further work, including developing principled acquisition functions for selecting the duration of long-run trials and improved multi-task models.










\bibliographystyle{ACM-Reference-Format}
\balance
\bibliography{references}

\appendix

\section{Experiment details}

\subsection{Time-varying effect transformation}
\label{app:tv_effect}
\begin{figure}[thb]
    \centering
    \includegraphics[width=0.25\textwidth]{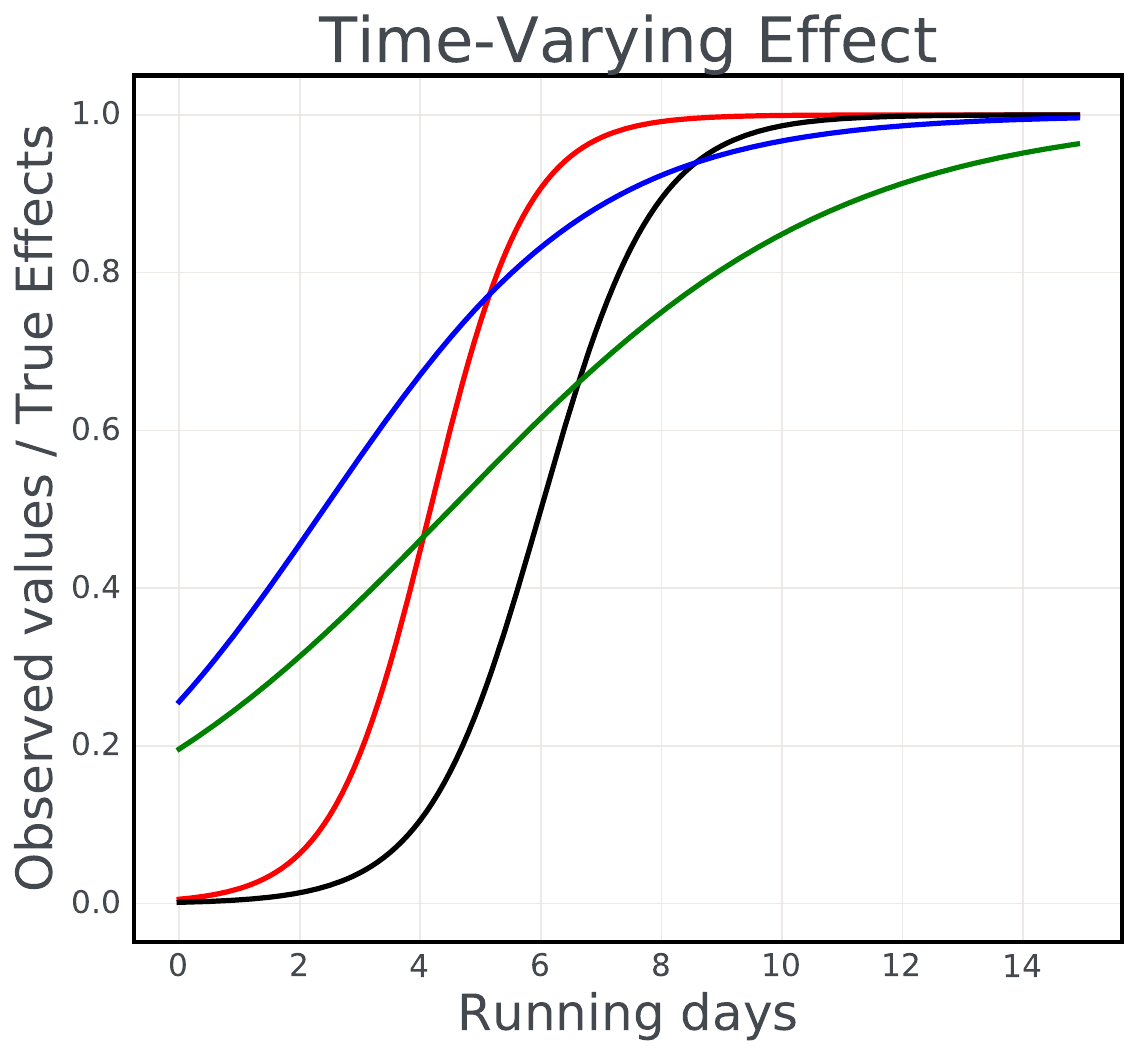}
        \vspace{-1.5ex}
    \caption{\label{fig:tv_effect} Example of time-varying effects across 4 arms. Each arm exhibits a unique convergence pattern toward its true long-term value.}
        \vspace{-1.5ex}
\end{figure}

We construct a sigmoid-based function to simulate time-varying effects in online experiments:
\begin{equation} g(x, t) = \frac{1}{1 + \exp\left(\frac{-\left(\frac{2t-T}{T} + 0.8x_j\right)}{\min(0.05+0.5x_i, 0.5)}\right)}. \label{eq:tv_trans}
\end{equation}
This function takes running time and parameter values as inputs and is monotonically increasing with respect to running time, simulating convergence to the true values over time. In the formula, $T$ represents the time needed to approach the true value, and $\frac{2t-T}{T}$ normalizes the running time $t$ with respect to the convergence time, signifying the current stage; the variables $x_i$ and $x_j$ are selected input dimensions that influence the convergence speed. Fig.~\ref{fig:tv_effect} depicts the progression of observed values relative to true long-term effects over the course of running time for four distinct arms in the Hartmann 3D experiment. As can be seen there, each arm exhibits a unique convergence pattern toward its true long-term value.

\subsection{Synthetic Functions}
\label{app:synth}
\begin{figure}[htb]
\centering
 \includegraphics[width=0.445\textwidth]{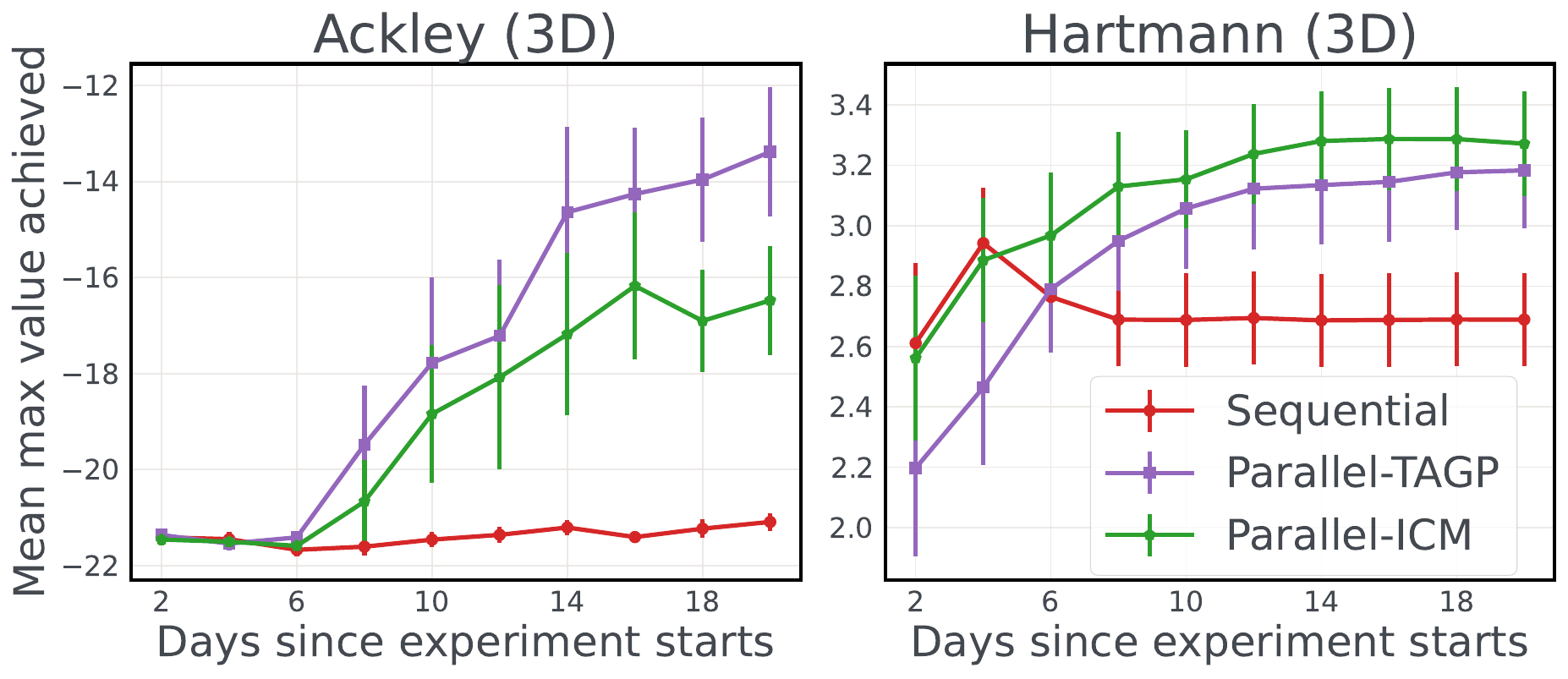}
 \vspace{-1.5ex}
\caption{Synthetic benchmark results. The parallel design outperforms sequential BO, highlighting the efficacy of our method in mitigating bias within short-term observations during rapid iterations.}
\label{fig:synthetic_result}
\vspace{-1.5ex}
\end{figure}

We considered two additional synthetic problems: the 3D Ackley function and the 3D Hartmann function. The time-varying transformation ~\ref{eq:tv_trans} was applied to each function to derive short-term measurements. In both scenarios, we set $i=0$ and $j=1$, utilizing the first two input dimensions to regulate the intensity of time-varying effects. The total resources were 32 arms for Ackley 3D and 24 for Hartmann 3D. The results are shown in Fig.~\ref{fig:synthetic_result}. The parallel design improved significantly over the sequential design, underscoring the crucial role of observing long-term effects in mitigating the bias inherent in short-term observations. The target-aware GP (TAGP) performs comparably to the ICM model in the Hartmann 3D experiment, since no additional short-term proxies are available beyond the objective itself. However, TAGP outperforms ICM in the Ackley experiment, attributable to Ackley being a more erratic function. The non-stationarity in the Ackley function results in negative transfer when sharing lengthscales in the kernel, harming the relative performance of the ICM kernel. 

\subsection{Video Playback Controller Optimization}
\label{app:abr}
We use the simulator developed by~\citep{feng2020cbo}, which is based on the open-source Park environment and is adapted to use network condition traces from real-world video playback streaming environments. Specifically, it utilizes 1912 traces of video playback sessions from the Facebook Android app, estimating chunk sizes for each segment of every playback session. Although this benchmark problem is simulator-based, the network conditions data are entirely derived from real-world scenarios, and the only simulation is of the application of the specific ABR policy to these actual conditions.

In the evaluation, we used a total of 24 arms for all methods.
The ABR controller is parametrized by four parameters: $p_b$, $p_o$, $p_c$, and $p_w$, subject to optimization through BO, with further details about the ABR in \citep{feng2020cbo}. The objective is to maximize a utility function $u(\mathbf{x})$ defined as $\text{QoE}(\mathbf{x}) - 20 \times (h_s(\mathbf{x}) - 40)_{+} - 20 \times (h_{bc}(\mathbf{x}) - 15)_{+}$, where $\text{QoE}(\mathbf{x})$ is the Quality of experience (QoE). QoE comes from the literature on model predictive control for video streaming~\citep{yin2015mpc}, and is represented as a weighted sum of the quality of the video chunks $h_b(\mathbf{x})$, stall time $h_s(\mathbf{x})$, and inter-temporal variation in quality $h_{bc}(\mathbf{x})$. The utility here incorporates additional penalties to prevent excessively long stall times and excessive inter-temporal variation in quality, both of which could adversely impact the overall user experience.
In addition to the overall metrics, we obtain the context-level breakdowns described in \citep{feng2020cbo}.
The target-aware model utilizes all short-term measurements of these metric breakdowns and the utility itself to predict the true long-term value of $u(\mathbf{x})$.
We applied the transformation of (\ref{eq:tv_trans}) with $T=15$ for all metrics. The convergence speed was contingent on the first two parameter dimensions.

In addition to assessing optimization performance, we also evaluate the out-of-sample prediction accuracy of these models, shown in Table~\ref{table:ll}. 
\begin{table}[htb]
\begin{small}
\begin{sc}
\begin{tabular}{lcccc}
\toprule
 & Sequential & ICM & Target-aware & \\
 \midrule
Video playback utility & -298275 &  \bf{-831} &  -952  \\
Retrieval utility & -22799742 & -461 &  \bf{-456}  \\
\bottomrule
\vspace{-1.5ex}
\end{tabular}
\end{sc}
\end{small}
\caption{Log-likelihood of out-of-sample predictions.}
\vspace{-4.5ex}
\label{table:ll}
\end{table}

\subsection{Retrieval Policy Optimization}
\label{app:sourcing}
This benchmark problem is designed to mimic a typical setup for retrieval in ranking systems at Meta. Although it is a toy model, it is based on real-world recommendation systems and captures the essence of a common data generation process in which recommendation engines produce topically related content and are then merged at subsequent stages. The simulator replicates the quality and infrastructure load of recommendations generated by a specific sourcing policy. The simulation setup was designed for research on sparse BO \citep{liu2023sparse}.
The sourcing system is modeled as a topic model, where each source possesses a distinct distribution over topics, and topics exhibit varying levels of relevance to the user. Similar sources may obtain duplicate items when they share topical similarities, which does not contribute to any improvement in recommendation quality.
The recommender sourcing system comprises 25 content sources, and we select 6 sources to consider a 6-dimensional retrieval policy $\mathbf{x}$ over the integer domain $[0, 50]^{6}$. In this context, each parameter specifies the number of items retrieved from a particular source, while the remaining sources have a fixed number of items to retrieve.

The objective is to maximize a utility function $u(\mathbf{x})$ defined as $\text{quality}(\mathbf{x}) - 0.6 \times \text{cost}(\mathbf{x}) - 20 \times (\text{cost}(\mathbf{x}) - 16)_{+}.$ The long-term objective is given by a noisy piece-wise utility function that incorporates both content relevance and infrastructure load. 
The overall content relevance score, denoted as $\text{quality}(\mathbf{x})$, is represented by the sum of content relevance scores after de-duplicating the retrieved content; $\text{cost}(\mathbf{x})$ stands for the infrastructure load, calculated as the sum of products of the number of retrievals and the cost per fetched item for each source.  The cost per fetched item varies across sources and is positively correlated with the source relevance score.  The utility function includes a penalty for high infrastructure cost. In a real-world scenario, infrastructure cost is budgeted for long-term system development. Following the same methodology as the video playback experiment, we apply the time-varying transformation to each metric with $T=15$ and convergence speed depending on the first two parameter dimensions. The evaluation uses a total of 32 arms for all methods.

In addition to the overall content relevance score, we calculate the quality score of each source, which is optimized by adjusting the number of retrieved items. These scores are weakly correlated with the overall value, as they depend on the duplications in the batching stage of sourcing. We leverage these scores in the target-aware model.  Table~\ref{table:ll} demonstrates that the Target-aware GP model outperforms other models through a robust approach that effectively utilizes multiple proxies. 

\subsection{7-d Ranking Policy Optimization}
In this problem, we utilize a large dataset from a real-world ranking policy optimization problem, evaluated across 8 distinct outcome metrics. This dataset is generated by an offline simulator that replays real-world traffic, capturing both short-term and hourly effects, as noted by \citep{letham2019jmlr} regarding the correlation between offline replay metrics and real-world online metrics. The simulator itself is a surrogate function fitted to 800 data points, which provides a relatively dense sampling compared to typical surrogate-based benchmarking literature \citep{eggensperger2015efficient}, allowing for accurate benchmarking based on real-world scenarios.

The total allocated resources for the the policies are 32. 
The same time-varying transformations are applied to derive short-term treatment effects. The objective is to maximize a piecewise utility function, constructed based on 4 key engagement metrics. This utility function aims to enhance a weighted sum of a specific subset of metrics $f_i(\mathbf{x})$ while penalizing the regression of the remaining ones, defined as
\begin{equation*}
\begin{aligned}
    u(\mathbf{x}) &= f_1(\mathbf{x}) + 0.5 \times f_2(\mathbf{x}) + 20\times(f_2(\mathbf{x}) + 0.5)_{-} + 0.5 \times f_3(\mathbf{x}) \\
         &\quad + 20 \times (f_3(\mathbf{x}) + 0.5)_{-} + 15 \times (f_4(\mathbf{x}) + 1.25)_{-}.
\end{aligned}
\end{equation*}
The utility function is geared towards optimizing the overall ranking policy for improved long-term outcomes.

\subsection{10-d Ranking Policy Optimization}
For this problem, we leverage data from a real-world experiment that spanned 16 days and involved 582 arms, each applied as treatment for a 4-hour period, with batches of 6 treatments running simultaneously. This extensive dataset provides a rich source of information that closely mimics real-world conditions. As with the other problem, we converted the real dataset into a benchmark problem by fitting a surrogate function to each metric.

To create a surrogate without time-of-day effects, we use as observations the time-averaged short-term treatment effect estimated with the surrogate. Additionally, we apply the same time-varying transformations to obtain both long-term and short-term effects, with the convergence time set to 15. The total allocated resources for the the policies are 48. The goal is to maximize a piecewise utility function, constructed based on three engagement metrics:
$$u(\mathbf{x}) = f_1(\mathbf{x}) - 0.5 \times f_2(\mathbf{x}) + 30\times(f_2(\mathbf{x})+1.0)_{-} + 20 \times (f_3(\mathbf{x}) - 0.5)_{+}.$$

\subsubsection{Optimization under Time-of-Day Effects}
\begin{figure}[t]
    \centering
    \includegraphics[width=0.28\textwidth]{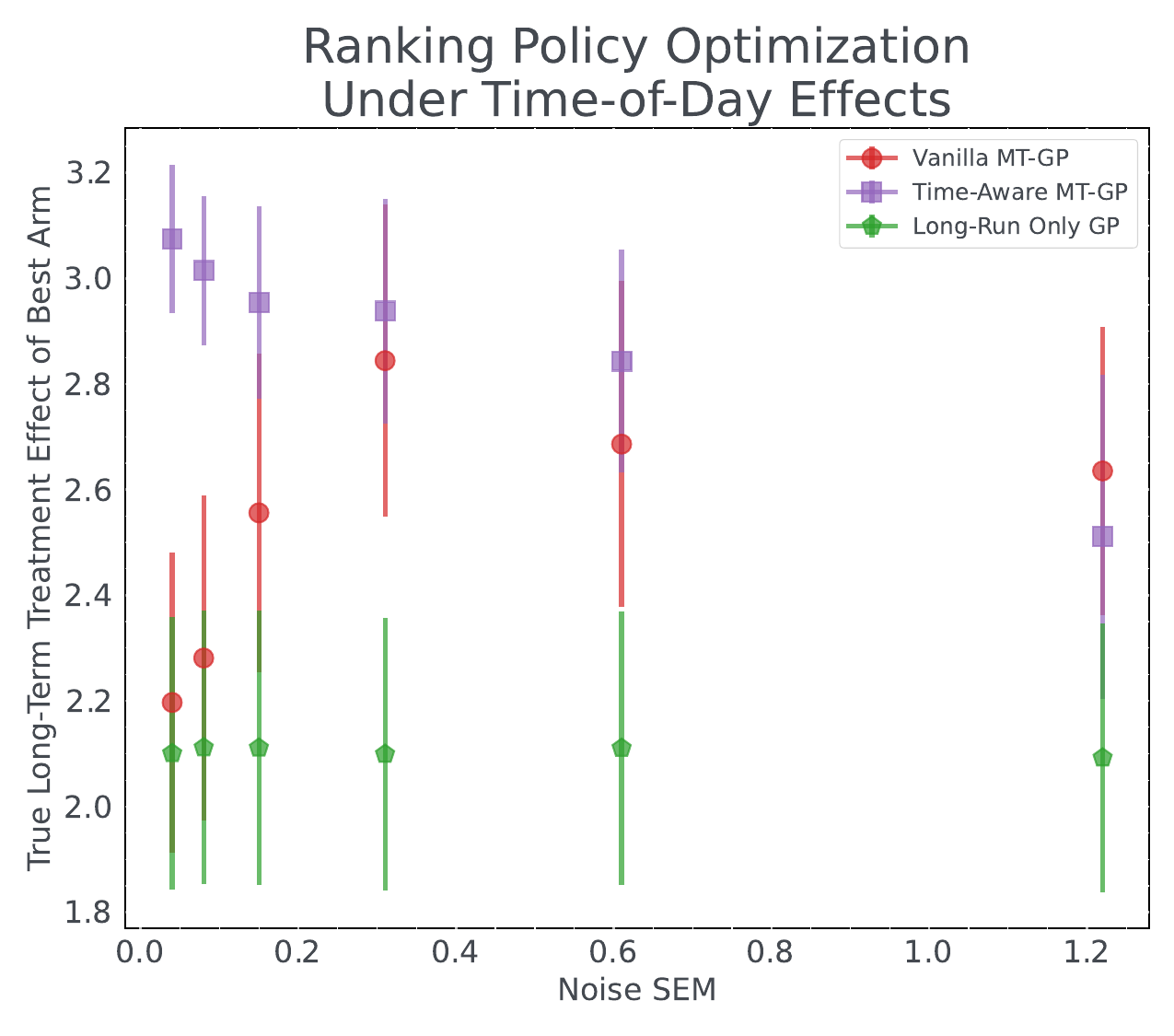}
    \vspace{-1.5ex}
    \caption{\label{fig:realtime_noise} A sensitivity analysis for optimizing the long-term treatment effect under time-of-day effects.}
    \vspace{-1.5ex}
\end{figure}

For the benchmark problem with time-of-day effects, we use the same surrogate to produce observations with time-of-day effects. The benchmark with time-of-day effects in the main text does not involve observation noise. However, in Fig.~\ref{fig:realtime_noise}, we evaluate the performance of various methods with respect the true long-term treatment effect for the final model-selected best point (out-of-sample) under various observation noise levels. The specified noise SEM is applied to the short-run observations and the 1/5 of the noise SEM is applied to the long run observations, which consistent with  empirical noise levels since the 4-hour short run trials are often run at low allocation since many observations are obtained.

At each iteration, the model-predicted time-averaged short-term effects and the long-term observations up to that point in time are used to fit a MTGP with an ICM kernel. 
Then, we optimize the posterior mean of that model for the long-run task to find the best out-of-sample arm, which is evaluated with respect to the true long-term effect (after 7-days).

\section{Time-aware Model Cross Validation}
\begin{figure}[h]
    \centering
    \includegraphics[width=0.47\textwidth]{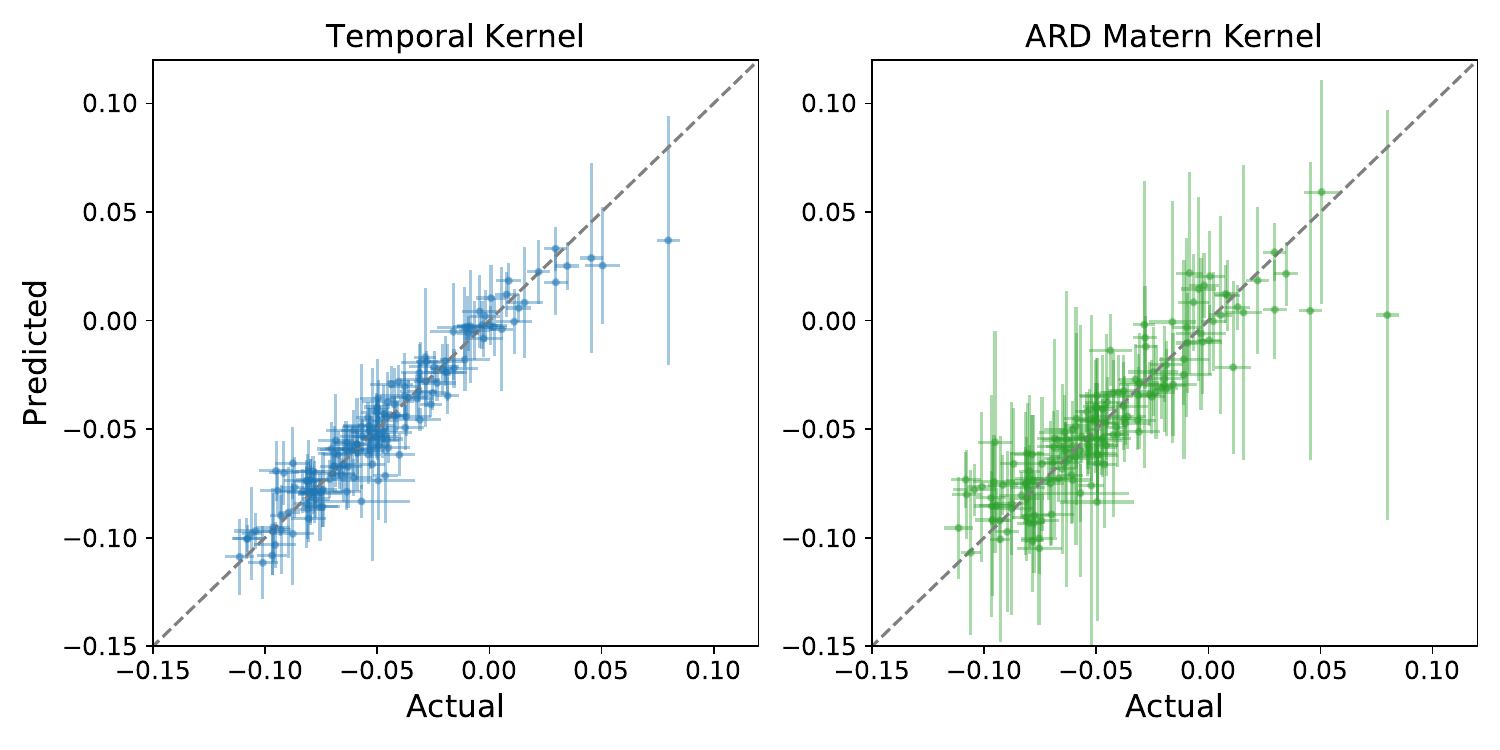}
    \vspace{-1.5ex}
    \caption{\label{fig:temp_vs_vanilla} Leave-one-out cross validation plots for the GP with a temporal kernel (left) and an ARD Mat\'ern 5/2 kernel (right). The temporal kernel significantly improves model fits on the engagement metric in short-run experiments.}
    \vspace{-1.5ex}
\end{figure}

\begin{table}[hbt]
\centering
\caption{\label{table:time_aware_model_diag} A comparison of a GP using the temporal kernel and a GP that uses a Mat\'ern kernel over the 11 tunable parameters and does not account for time of day. We report mean squared error (MSE) and negative log-likelihood (NLL) (computed using leave-on-out cross-validation) on 13 engagement metrics for which noisy observations were collected during an online experiment on Instagram.
}
\begin{small}
\begin{sc}
\begin{tabular}{lcccc}
\toprule
 & Temporal & Mat\'ern & Temporal & Mat\'ern\\
 \midrule
  &  \multicolumn{2}{c}{NLL} &  \multicolumn{2}{c}{MSE}\\
  \midrule
Metric 1 &\bf{470.34} & 606.57 & 1.89 & \bf{1.38} \\
Metric 2 &\bf{1015.9} & 1453.58 & \bf{15.15} & 15.22 \\
Metric 3 &\bf{903.06} & 912.08 & \bf{8.71} & 9.68 \\
Metric 4 &\bf{299.02} & 486.94 & \bf{0.14} & 0.15 \\
Metric 5 &\bf{1435.37} & 2255.96 & \bf{16.22} & 16.24 \\
Metric 6 &400.46 & \bf{375.69} & 1.2 & \bf{0.93} \\
Metric 7 &\bf{1112.11} & 1243.23 & 11.13 & \bf{10.27} \\
Metric 8 &\bf{384.69} & 416.78 & 0.83 & \bf{0.78} \\
Metric 9 &\bf{793.25} & 1145.56 & \bf{4.15} & 4.2 \\
Metric 10 &\bf{1399.76} & 1423.12 & \bf{5.81} & 6.57 \\
Metric 11 &\bf{251.63} & 338.19 & 0.39 & \bf{0.31} \\
Metric 12 &\bf{265.04} & 362.83 & 0.34 & \bf{0.26} \\
Metric 13 &\bf{346.67} & 413.12 & 0.96 & \bf{0.59} \\
\bottomrule
\end{tabular}
\end{sc}
\end{small}
\vspace{-2ex}
\end{table}

\end{document}